\documentclass[runningheads]{llncs}
\usepackage{graphicx}
\usepackage{amsmath,amssymb} 

\usepackage{color}

\begin{document}
\title{Decoding Generic Visual Representations From Human Brain Activity using Machine Learning}

\titlerunning{ }

\author{Angeliki Papadimitriou \and
Nikolaos Passalis \and
Anastasios Tefas}

\authorrunning{A. Papadimitriou, N. Passalis and A. Tefas}

\institute{Artificial Intelligence and Information Analysis Laboratory\\ Aristotle University of Thessaloniki, Thessaloniki 541 24, Greece}
\maketitle              
\begin{abstract}
Among the most impressive recent applications of neural decoding is the {visual representation decoding}, where the category of an object that a subject either sees or imagines is inferred by observing his/her brain activity. Even though there is an increasing interest in the aforementioned visual representation decoding task, there is no extensive study of the effect of using different machine learning models on the decoding accuracy. In this paper we provide an extensive evaluation of several machine learning models, along with different similarity metrics, for the aforementioned task, drawing many interesting conclusions. That way, this paper a) paves the way for developing more advanced and accurate methods and b) provides an extensive and easily reproducible baseline for the aforementioned decoding task.

\end{abstract}
\section{Introduction}
Neural decoding is the process of deciphering human brain activity of subjects performing a specific task.  In order to record a subject's brain activity, several functional neuroimaging techniques are available. The term ``functional'' refers to the temporal aspect that allows these methods to capture the changes of brain activity over time. Such methods are Electroencephalography (EEG)~\cite{niedermeyer2005electroencephalography} and functional Magnetic Resonance Imaging (fMRI)~\cite{huettel2004functional}, which have been extensively employed in the literature to study human brain activity during various tasks, such as sleep~\cite{tagliazucchi2014decoding,horikawa2013neural}, walking~\cite{gwin2010removal,presacco2011neural}, and dancing~\cite{fink2009brain,cruz2014neural}.

Among the most impressive applications of fMRI is the \textit{visual representation decoding}~\cite{horikawa2017generic}. Visual representation decoding refers to the \textit{prediction} of what a subject sees or imagines by observing \textit{only} his/her neural activity. More specifically, the subject encodes the stimulus as an internal neural representation and the goal is to design a model that can recover that  representation by partially observing the brain state (e.g., by observing  EEG or  fMRI signals) in order to understand or even reconstruct the stimulus that evoked it~\cite{horikawa2017generic,wen2017neural,guccluturk2017reconstructing,wen2018transferring}. 

However, it is not possible to infer the actual internal neural representations of a subject and use them to train models that can perform the aforementioned decoding task. To overcome this limitation, the representation must be supplied from an external source. In~\cite{horikawa2017generic} it has been shown that the features extracted from various levels of a deep Convolutional Neural Network (CNN) are tightly correlated with the brain activity observed from various brain regions. Therefore, usually the features extracted from a CNN are employed as the intermediate neural representation that can be used to decode what a subject actually sees or imagines~\cite{horikawa2017generic,wen2018transferring}. Then, the visual representation can be decoded from the observed brain signals using machine learning models~\cite{nasrabadi2007pattern}. Even though there is an increasing interest in the aforementioned visual representation decoding task, there is no extensive study of the effect of using different machine learning models, as well as of the effect of various hyper-parameters employed during this process, on the decoding accuracy.

The main contribution of this paper is the extensive evaluation of different machine learning models, along with different similarity metrics, for the recently proposed task of decoding visual representations from human brain activity signals. That way, this paper a) paves the way for developing more advanced and accurate decoding methods and b) provides an extensive and easily reproducible baseline for the aforementioned task. To this end, all the experiments performed in this paper can be readily reproduced using the code available on-line at \url{https://github.com/angpapadi/Visual-Representation-Decoding-from-Human-Brain-Activity}.

Several interesting conclusions are drawn from the results presented in this paper. Even though different models seem to behave differently for the various decoding tasks, using a Multilayer Perceptron (MLP) seems to provide the best decoding accuracy. The choice of the similarity measure for the decoding process can also be of crucial importance for  some tasks. Furthermore, several conclusions regarding the actual way that the brain works can be also drawn from the results, e.g., in imagery tasks brain regions that belong to the higher visual cortex seem to exhibit higher predictive power, while lower visual cortex regions seem to encode lower level features that require the use of non-linear models for the decoding, etc.

The rest of this paper is structured as follows. The used decoding method, along with the evaluated machine learning models and similarity metrics are described in Section~\ref{section:proposed}, while the experimental protocol and evaluation results are provided in Section~\ref{section:experiments}. Finally, conclusions are drawn and future research directions are discussed in Section~\ref{section:conclusions}.

\section{Machine Learning for Visual Representations Decoding}
\label{section:proposed}

\begin{figure}
\centering
\includegraphics[width=0.99\linewidth]{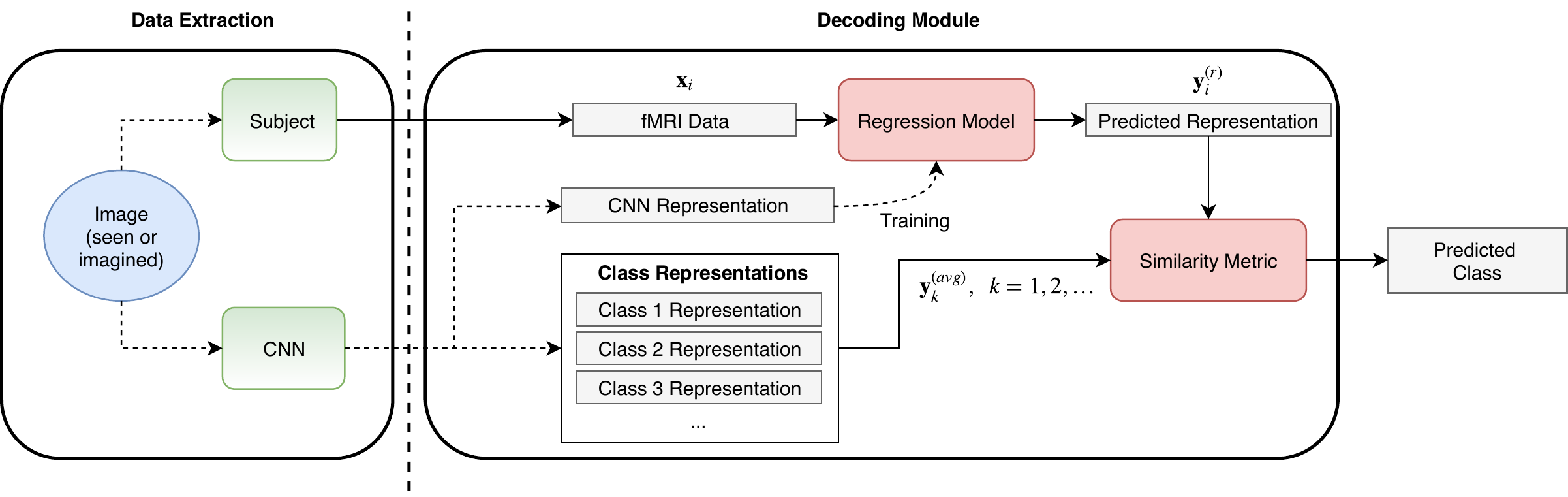}
\caption{Pipeline for decoding generic visual representations from human brain activity }
\label{fig:pipeline}
\end{figure}

The method used in this paper builds upon the generic brain activity decoding pipeline proposed in~\cite{horikawa2017generic}. Fig.~\ref{fig:pipeline} provides an overview of the employed decoding pipeline. First, a subject either views an image or is instructed to imagine a concept (related to a category of images viewed before), while his/her brain activity is monitored using fMRI. Also, a CNN is used to extract a feature representation from the corresponding images (as they are presented to the subject). The use of a CNN layer for this task is not arbitrary. As discussed in~\cite{horikawa2017generic}, the features extracted from various layers of a CNN are tightly associated with the activity of various part of the human visual cortex. This allows for using these representations as an intermediate step for decoding the human brain activity for specific visual tasks, i.e., recognizing the class of an image shown to or imagined by a subject. To this end, a regression model is trained to  predict the representation of the viewed image directly using the measured fMRI signals. Then, the class of the image can be inferred by measuring the similarity between the predicted representation to a set of prototype class representations, where each class is represented by the average feature vector extracted by feeding a set of images that belong to a specific class to the used CNN. Note that by regressing the image representation, instead of merely classifying the brain signals into a set of predetermined classes, allows for using the aforementioned pipeline for image categories that were \textit{never presented} to the subject and/or regression model during the training.

The decoding pipeline can be more formally defined as follows. Let $\mathbf{x}_i \in \mathbb{R}^D$ denote a $D$-dimensional feature vector appropriately extracted from the measured fMRI signals when the $i$-th experiment is performed~\cite{horikawa2017generic}. Also, let $\mathbf{y}^{(t)}_{i} \in \mathbb{R}^L$ be the representation extracted from a layer of a CNN, when the image used for the $i$-th experiment is fed to the network, and $L$ denote the dimensionality of the extracted feature representation. The prototype class representation for the $k$-th class is defined as: $\mathbf{y}^{(avg)}_{k} = \frac{1}{|\mathcal{R}_k|} \sum_{\mathbf{y} \in \mathcal{R}_k} \mathbf{y}$, where $\mathcal{R}_k$ is the set of CNN representations extracted from the images that belong to class $k$ and $|\mathcal{R}_k|$ is the cardinality of this set. Then, a machine learning model $f_\mathbf{W}(\mathbf{x})$ is used to regress the aforementioned image representation, where $\mathbf{W}$ denotes the parameters of the employed model. The  output of this model is denoted by $\mathbf{y}^{(r)}_i = f_\mathbf{W}(\mathbf{x}_i) \in \mathbb{R}^L$ and can be used to infer the class of the corresponding image by measuring the similarity of $\mathbf{y}^{(r)}_i$ with each of the class representations $\mathbf{y}^{(avg)}_{k}$. Therefore, the predicted class $k^*$ of the object  the subject sees or imagines is calculated as:
\begin{equation}
k^* = \arg \max_{k} S(\mathbf{y}^{(r)}_i, \mathbf{y}^{(avg)}_{k})
\end{equation}
where $S(\mathbf{a}, \mathbf{b})$ is an appropriately defined similarity metric between two vectors $\mathbf{a} \in \mathbb{R}^L$ and $\mathbf{b} \in \mathbb{R}^L$. Note that the CNN representation $\mathbf{y}^{(r)}_i$ is only needed during the training process of the regression model, as also shown in Fig.~\ref{fig:pipeline}. Then, the class of the object that a subject sees or imagines can be inferred (during the test) without having access to the corresponding image representation, since only the (precomputed) class representation vectors are needed.

As already stated in Section 1, the main contribution of this paper is the extensive evaluation of various machine learning models for the task of decoding generic visual representations from human brain activity. To this end, four different regression models are evaluated in this paper. The used models are briefly described bellow:
\begin{enumerate}
\item \textbf{k-Nearest Neighbor Regression (kNN):} In k-Nearest Neighbor Regression the $k$ nearest neighbors of a data sample are used to regress its target by averaging the (known) target feature vectors of its $k$ neighbors~\cite{hastie1996discriminant}. The contribution of each neighbor is appropriately weighted according to its distance to the current test sample.
\item \textbf{Linear Regression (LR):} In Linear Regression the output of the model is calculated as $f_\mathbf{W}(\mathbf{x}) = \mathbf{W}_{lr}\mathbf{x} + \mathbf{b}_{lr}$~\cite{mosteller1977data}, where $\mathbf{W}_{lr}\ \in \mathbb{R}^{L\times D}$ is a matrix that contains the regression parameters and $\mathbf{b}_{lr} \in \mathbb{R}^{L}$ is the independent term in the linear model. The model is trained to minimize the mean squared error between its output and the target representation given by:
\begin{equation}
\label{eq:mse}
\mathcal{L}_{mse} = \frac{1}{2N} \sum_{i} ||f_\mathbf{W}(\mathbf{x_i}) - \mathbf{y}^{(t)}_{i}||_2^2,
\end{equation}
where $||\mathbf{x}||_2$ denotes the $l^2$ norm of a vector $\mathbf{x}$ and $N$ is the number of data samples used to fit the regression model. The model can be \textit{regularized} to avoid overfitting, e.g., using the $l^2$ norm of the regression coefficients as regularizer leads to Ridge Regression (RR)~\cite{hoerl1970ridge}.
\item \textbf{Kernel Regression (KR):} Kernel Regression is a non-linear variant of LR, where the data are first (non-linearly) projected into a higher-dimensional space, where they can be better separated~\cite{shawe2004kernel}. KR employs the so-called \textit{kernel trick} to allow for efficiently fitting the regression model, even when the data are projected into an infinite dimensional space.

\item \textbf{Multilayer Perceptrons (MLP):} Multilayer Perceptrons are powerful neural network-based methods that can model complex relationships between the input data and their targets through multiple non-linear layers~\cite{haykin2009neural}. Several methods have been proposed for training and designing MLP networks. In this work, the networks are trained to minimize the loss function given in (\ref{eq:mse}) using the Adam optimizer~\cite{kingma2014adam}. MLPs are often prone to overfitting the data, especially when a small number of training samples is used. To overcome this issue, regularization methods, such as Dropout~\cite{srivastava2014dropout} (also abbreviated as ``drop.''), can be used.
\end{enumerate}
The selected hyper-parameters for the evaluated models, e.g., number of hidden units in the MLP, regularizer weight, etc., are presented in Section~\ref{section:experiments}. Both the input vectors $\mathbf{x}_i$, as well as the extracted representations $\mathbf{y}^{(t)}_i$, are normalized to have zero mean and unit variance (z-score normalization). The output of the regression model is then appropriately ``denormalized'', using the mean and standard deviation computed on the train set, in order to compute the similarity with the prototype class representation vectors $\mathbf{y}^{(avg)}_{k}$.

Apart from the model used for the regression process, the accuracy of the proposed  decoding pipeline critically relies on employed similarity metric.  Three different similarity metrics $S(\mathbf{a}, \mathbf{b})$ are evaluated in this paper:
\begin{enumerate}
\item \textbf{Euclidean similarity:} The Euclidean similarity is computed as the inverse of Euclidean distance: $S_{euclidean}(\mathbf{a}, \mathbf{b}) = \frac{1}{1 + ||\mathbf{a}-\mathbf{b}||_2}$.
\item \textbf{Cosine similarity:} The cosine similarity is defined as the angle between two vectors: $S_{cosine}(\mathbf{a}, \mathbf{b}) = \frac{\mathbf{a}^T\mathbf{b}}{||\mathbf{a}||_2 ||\mathbf{b}||_2 }$.
\item \textbf{Pearson similarity:}  The Pearson similarity (correlation) between two vectors is computed as: $S_{pearson}(\mathbf{a}, \mathbf{b}) = \frac{(\mathbf{a} - {\mu}_a)^T (\mathbf{b} - {\mu}_a)}{||\mathbf{a} - {\mu}_a||_2 ||\mathbf{b} - {\mu}_a||_2}$, where ${\mu}_a$ and ${\mu}_b$ are the average of the values in vectors $\mathbf{a}$ and $\mathbf{b}$ respectively.
\end{enumerate}

\section{Experimental Evaluation}
\label{section:experiments}

In this paper the dataset provided by the authors of~\cite{horikawa2017generic} is used, while we also closely follow the experimental setup and evaluation metrics described in their work. More specifically, the dataset contains fMRI data from 5 subjects, where each subject was presented with 1,200 images from 150 object categories to form the train set and 50 additional images from 50 different categories were presented to them to form the test set. It should be stressed that different image categories are contained in the train set and test set, prohibiting the use of traditional classification algorithms for the task of neural decoding. Also, two different types of experiments were conducted: a) image presentation experiments, where the aforementioned images were presented to the subjects, and b) imagery experiments, where the subjects were instructed to imagine one image that belongs to one of the 50 categories of the test set. The dataset contains a total of 1,750 test samples for the image presentation experiments and 500 test samples per subject for the imagery experiments. The interested reader is referred to~\cite{horikawa2017generic} for more details regarding the data collection protocol and experimental setups. Furthermore, for all the conducted experiments, the preprocessed fMRI and CNN feature vectors provided by the authors of~\cite{horikawa2017generic} were used. Regarding the CNN architecture, an AlexNet with 5 convolutional (CNN1-CNN5) and 3 fully connected layers (FC1-FC3), trained on the Imagenet dataset was used~\cite{krizhevsky2012imagenet}. From each layer 1,000 random units were sampled and used as the extracted feature representation (before applying the activation function). The feature vectors provided by~\cite{horikawa2017generic} are used, to allow for easily reproducing the conducted experiments.  

The hyper-parameters selected for the evaluated models are briefly described bellow. For the k-Nearest Neighbor model the $k=5$ nearest neighbors were used. For the Ridge Regression the weight of the regularizer was set to 1, while for the Kernel Regression several different kernels were evaluated. The best results were obtained when a 2nd degree polynomial kernel was used and the regression model was regularized (Kernel Ridge Regression)~\cite{vovk2013kernel}. The weight of the regularizer was set to 0.005, while the constant term of the kernel was set to 10. Finally, an MLP with one hidden layer with 300 neurons with sigmoid activations was chosen, after experimenting with many different MLP architectures/activation functions and evaluating their regression accuracy. The MLP was also combined with the Dropout method (dropout probability 30\% for the input layer). All the MLP models were trained for 100 epochs (batch size 128) with a learning rate of 0.001 (the default hyper-parameters for the Adam algorithm were used). The LR, KR and kNN models were implemented using the scikit-learn library~\cite{scikit-learn}, while the MLP models were implemented using the keras library~\cite{chollet2015keras}. 

The evaluation results for the image presentation experiments are reported in Table~\ref{table:image-present-eval}. The models were trained to regress the features vectors extracted from the last (CNN5) convolutional layer of the CNN, since this layer has been shown to provide the best neural decoding accuracy~\cite{horikawa2017generic}. The features extracted from the CNN are decoded using different regions selected from the fMRI data:  the lower visual cortex areas (V1-V4) and higher visual cortex areas, namely the lateral occipital complex (LOC), fusiform face area (FFA) and parahippocampal place area (PPA). Voxels from regions V1-V3 are combined to form the lower visual cortex (LVC) region and voxels from LOC, FFA and PPA form the higher visual cortex (HVC). The whole visual cortex is denoted by ``VC''. The Pearson similarity is used in the conducted experiments. All the extracted voxels from the corresponding brain regions were used in the conducted experiments, i.e., we did not select only the voxels with the highest correlation, as in~\cite{horikawa2017generic}. This allows for retaining as much information as possible. The decoding accuracy is measured as proposed in~\cite{horikawa2017generic}: every possible combination of the correct class against all the classes is considered and the mean decoding accuracy is reported. Note that a random classifier will achieve a decoding accuracy of 50\% under this binary evaluation setup. 

Several interesting conclusions can be drawn from the results reported in Table~\ref{table:image-present-eval}. First, the non-linear models (KR, MLP) achieve significantly higher decoding accuracy for the lower visual cortex regions (V1-V4, LVC), as well as for the LOC and FFA regions.  For example the decoding accuracy from the V2 region increases from 76.29\% to 84.68\%, when an MLP is used instead of a LR model. This possibly suggests that complex non-linear relationships between the extracted voxels and the corresponding deep features indeed exist for these (individual) regions. On the other hand, when voxels from the whole visual cortex are used as input (VC), then all the evaluated models (except for the kNN) achieve  almost the same decoding accuracy, ranging from 93.16\% to 93.95\%. Also, the differences between the predictive power of the LVC and HVC regions are small. For example, the MLP achieves 89.74\% decoding accuracy when using the features extracted from the LVC region and 90.14\% decoding accuracy when using the features extracted from the HVC region, demonstrating that can reliably decode the image representations regardless the used region.  Using regularization seems to slightly improve the decoding accuracy of linear regression. However, this is not true for the MLP, which achieves the best decoding accuracy without any form of regularization (the differences are small though). Using a non-linear regressor (KR) seems to significantly improve the results over the plain linear regression in most of the cases. However, the kNN model seems to be unable to reliably decode the features. This can be possibly attributed to the highly non-linear nature of the model that mostly relies on the local properties of the input space that prohibits the effective generalization for samples that do not follow the original distribution. Finally, note that the MLP achieves the best decoding accuracy, significantly outperforming the rest of the models.

\begin{table}
\begin{center}
\caption{Image Presentation Evaluation: Decoding accuracy for different machine learning models}
\label{table:image-present-eval}
\begin{tabular}{l|cccc|c|ccc|c|c}
Model & V1 & V2 & V3 & V4 & LVC & LOC & FFA & PPA & HVC & VC \\
\hline
LR &77.49&76.29&83.51&88.55&86.47&83.25&81.93&84.01&88.67&93.16\\
RR &79.05&78.92&84.47&88.85&86.81&83.82&82.52&\underline{84.03}&88.83&93.17\\
\hline
kNN &74.58&73.97&72.94&75.82&75.12&75.17&75.33&74.66&76.29&77.70\\
KR &80.64 &83.32 & \underline{87.14} & \underline{89.06} & \underline{89.63} & 85.33 &84.77 &82.71 &89.81 &93.45\\
MLP & \underline{81.95}  & \textbf{84.68 }& \textbf{87.73}& \textbf{90.68}& \textbf{89.74}& \textbf{86.47}& \textbf{86.13} & \textbf{84.29} &\textbf{90.14} & {\textbf{93.95}}  \\
MLP (drop.) & \textbf{82.10} & \underline{83.56} & 86.54 & 88.90 & 88.73 & \underline{85.73} & \underline{85.10} & 82.92 & \underline{90.01} & \underline{93.75} \\
\end{tabular}
\end{center}
\end{table}

The corresponding results for the imagery experiments are reported in Table~\ref{table:imagery-eval}. For the imagery experiments it seems that the HVC region provides significantly better decoding accuracy than the LCV region. For example, the MLP using dropout achieves 71.04\% decoding accuracy when using the HVC features as input, while the accuracy drops to 63.63\% when using the LVC as input. This can be possibly attributed to the lack of physical stimulus for the imagery evaluation, and can possibly hint that the imagery tasks are mostly related to the areas of the higher visual cortex. Even though for the imagery tasks the MLP is the second best performing model, the kNN achieves the best decoding accuracy for almost every brain region. This highlights the importance of using a machine learning model that fits the problem at hand, i.e., the imagery-based decoding is significantly harder than the image presentation-based decoding, requiring significantly more powerful and non-linear models. Nonetheless, it should be stressed that MLPs (when combined with the appropriate regularizer) seem to be able to handle this task especially well, compared to the rest of the models.

\begin{table}
\begin{center}
\caption{Imagery Evaluation: Decoding accuracy for different machine learning models}
\label{table:imagery-eval}
\begin{tabular}{l|cccc|c|ccc|c|c}
Model & V1 & V2 & V3 & V4 & LVC & LOC & FFA & PPA & HVC & VC\\
\hline
LR &59.45&57.36&62.91&64.94&60.2&61.87&61.41&68.21&62.87&65.19\\
RR &59.32&58.4&63.15&65.32&60.64&62.12&61.81&68.26&63.27&65.18\\
\hline
kNN &\textbf{67.87}&\textbf{66.68}&\textbf{67.49}&\textbf{70.18}&\textbf{67.51}&\textbf{70.66}&\textbf{69.62}&69.32&\underline{70.63}&\textbf{69.99}\\
KR &57.29&59.2&65.1&68.28&61.37&65.06&63.27&66.63&65.32&65.00\\
MLP & 60.20 & \underline{63.98} & 66.07 & 68.93 & 63.45 & 69.24 & 68.33 & \textbf{70.39} & 67.89 & 65.73 \\
MLP (drop.) & \underline{62.50} & 62.05 & \underline{66.74} & \underline{70.00} & \underline{63.63} & \underline{70.38} & \underline{69.26} & \underline{69.79} & {\textbf{71.04 }} & \underline{68.14} \\
\end{tabular}
\end{center}
\end{table}

In the previous experiments, the Pearson correlation was used as similarity metric. The effect of three different similarity metrics on the accuracy of the best model for the image presentation (MLP) and imagery experiments (knn) is evaluated in Table~\ref{table:similarity-eval}. For the image presentation experiments (denoted by ``(IP)'') the differences are quite small, even though the Pearson similarity achieves the best accuracy in most of the cases. On the other hand, the Euclidean similarity seems to perform significantly better than the other two similarity measures for the imagery task (denoted by ``(I)''). Again, this highlights the importance of using the appropriate model, along with an appropriately selected similarity metric, for tasks with different characteristics.

\begin{table}
\begin{center}
\caption{Evaluating the effect of different similarity metrics on the decoding accuracy}
\label{table:similarity-eval}
\begin{tabular}{l|cccc|c|ccc|c|c}

Model & V1 & V2 & V3 & V4 & LVC & LOC & FFA & PPA & HVC & VC \\
\hline
MLP+Eucl.(IP) & 81.86 & 83.78 & 86.17 & 88.30 & 88.13 & 84.79 & 84.42 & 82.87 & 88.11 & 92.47 \\
MLP+Cos.(IP) & \textbf{84.39} & 83.98 & 82.58 & 87.66 & 81.25 & 83.48 & 85.94 & \textbf{88.05} & 87.81 & 92.36 \\
MLP+Pear.(IP)   & {81.95}  & \textbf{84.68 }& \textbf{87.73}& \textbf{90.68}& \textbf{89.74}& \textbf{86.47}& \textbf{86.13} & {84.29} &\textbf{90.14} & {\textbf{93.95}}  \\
\hline
\hline
knn+Eucl.(I) & \textbf{71.38} & \textbf{70.57} & \textbf{70.98}& \textbf{72.85} & \textbf{71.01} &\textbf{73.11 }&\textbf{72.56} &\textbf{72.38} &\textbf{73.17 }&\textbf{72.75} \\
knn+Cos.(I) & 70.66 & 69.81 & 70.20 & 72.01 & 70.39 & 72.39 & 71.67 & 71.54 & 72.32 & 71.88 \\
knn+Pear.(I) & {67.87}&{66.68}&{67.49}&{70.18}&{67.51}&{70.66}&{69.62}&69.32&{70.63}&{69.99}\\
\hline
\end{tabular}
\end{center}
\end{table}

\section{Conclusions}
\label{section:conclusions}

In this paper an extensive evaluation of different machine learning models and similarity measures for the task of decoding deep visual representations from human brain activity signals was presented and several interesting conclusions were drawn. For example,  different models seem to behave differently for the various decoding tasks, while using a Multilayer Perceptron (MLP) seems to provide the best decoding accuracy for most of the tasks. Furthermore, the choice of an appropriate similarity measure was shown to be of crucial importance for  some tasks. Finally, observations regarding how the brain actually works can also be possibly deduced from the results reported in this paper. 

There are many interesting future research directions. Metric learning methods can be used to learn the optimal similarity measure, further increasing the decoding accuracy. Convolutional Neural Networks can be used to better model the spatial relationships between voxels~\cite{krizhevsky2012imagenet}, while Recurrent Neural Networks can be used to model the temporal behavior of brain activity~\cite{graves2013speech}. Finally, transfer learning~\cite{pan2010survey} and cross-modal knowledge transfer methods~\cite{passalis} can be employed to appropriately transfer the knowledge that can be shared among different subjects, leading to the development of effective subject-agnostic decoding models.


\begin{thebibliography}{10}

\bibitem{niedermeyer2005electroencephalography}
Niedermeyer, E., da~Silva, F.L.:
\newblock Electroencephalography: basic principles, clinical applications, and
  related fields.
\newblock Lippincott Williams \& Wilkins (2005)

\bibitem{huettel2004functional}
Huettel, S.A., Song, A.W., McCarthy, G.,  et~al.:
\newblock Functional magnetic resonance imaging. Volume~1.
\newblock Sinauer Associates Sunderland, MA (2004)

\bibitem{tagliazucchi2014decoding}
Tagliazucchi, E., Laufs, H.:
\newblock Decoding wakefulness levels from typical fmri resting-state data
  reveals reliable drifts between wakefulness and sleep.
\newblock Neuron \textbf{82}(3) (2014)  695--708

\bibitem{horikawa2013neural}
Horikawa, T., Tamaki, M., Miyawaki, Y., Kamitani, Y.:
\newblock Neural decoding of visual imagery during sleep.
\newblock Science \textbf{340}(6132) (2013)  639--642

\bibitem{gwin2010removal}
Gwin, J.T., Gramann, K., Makeig, S., Ferris, D.P.:
\newblock Removal of movement artifact from high-density eeg recorded during
  walking and running.
\newblock Journal of Neurophysiology \textbf{103}(6) (2010)  3526--3534

\bibitem{presacco2011neural}
Presacco, A., Goodman, R., Forrester, L., Contreras-Vidal, J.L.:
\newblock Neural decoding of treadmill walking from noninvasive
  electroencephalographic signals.
\newblock Journal of neurophysiology \textbf{106}(4) (2011)  1875--1887

\bibitem{fink2009brain}
Fink, A., Graif, B., Neubauer, A.C.:
\newblock Brain correlates underlying creative thinking: Eeg alpha activity in
  professional vs. novice dancers.
\newblock NeuroImage \textbf{46}(3) (2009)  854--862

\bibitem{cruz2014neural}
Cruz-Garza, J.G., Hernandez, Z.R., Nepaul, S., Bradley, K.K., Contreras-Vidal,
  J.L.:
\newblock Neural decoding of expressive human movement from scalp
  electroencephalography (eeg).
\newblock Frontiers in human neuroscience \textbf{8} (2014)  188

\bibitem{horikawa2017generic}
Horikawa, T., Kamitani, Y.:
\newblock Generic decoding of seen and imagined objects using hierarchical
  visual features.
\newblock Nature communications \textbf{8} (2017)  15037

\bibitem{wen2017neural}
Wen, H., Shi, J., Zhang, Y., Lu, K.H., Cao, J., Liu, Z.:
\newblock Neural encoding and decoding with deep learning for dynamic natural
  vision.
\newblock Cerebral Cortex  1--25

\bibitem{guccluturk2017reconstructing}
G{\"u}{\c{c}}l{\"u}t{\"u}rk, Y., G{\"u}{\c{c}}l{\"u}, U., Seeliger, K., Bosch,
  S., van Lier, R., van Gerven, M.A.:
\newblock Reconstructing perceived faces from brain activations with deep
  adversarial neural decoding.
\newblock In: Proceedings of the Advances in Neural Information Processing
  Systems. (2017)  4246--4257

\bibitem{wen2018transferring}
Wen, H., Shi, J., Chen, W., Liu, Z.:
\newblock Transferring and generalizing deep-learning-based neural encoding
  models across subjects.
\newblock NeuroImage \textbf{176} (2018)  152--163

\bibitem{nasrabadi2007pattern}
Nasrabadi, N.M.:
\newblock Pattern recognition and machine learning.
\newblock Journal of Electronic Imaging \textbf{16}(4) (2007)  049901

\bibitem{hastie1996discriminant}
Hastie, T., Tibshirani, R.:
\newblock Discriminant adaptive nearest neighbor classification and regression.
\newblock In: Proceedings of the Advances in Neural Information Processing
  Systems. (1996)  409--415

\bibitem{mosteller1977data}
Mosteller, F., Tukey, J.W.:
\newblock Data analysis and regression: a second course in statistics.
\newblock Addison-Wesley Series in Behavioral Science: Quantitative Methods
  (1977)

\bibitem{hoerl1970ridge}
Hoerl, A.E., Kennard, R.W.:
\newblock Ridge regression: Biased estimation for nonorthogonal problems.
\newblock Technometrics \textbf{12}(1) (1970)  55--67

\bibitem{shawe2004kernel}
Shawe-Taylor, J., Cristianini, N.,  et~al.:
\newblock Kernel methods for pattern analysis.
\newblock Cambridge university press (2004)

\bibitem{haykin2009neural}
Haykin, S.S.:
\newblock Neural networks and learning machines. Volume~3.
\newblock Pearson Education Upper Saddle River (2009)

\bibitem{kingma2014adam}
Kingma, D., Ba, J.:
\newblock Adam: A method for stochastic optimization.
\newblock In: Proceedings of the International Conference on Learning
  Representations. (2014)

\bibitem{srivastava2014dropout}
Srivastava, N., Hinton, G., Krizhevsky, A., Sutskever, I., Salakhutdinov, R.:
\newblock Dropout: a simple way to prevent neural networks from overfitting.
\newblock The Journal of Machine Learning Research \textbf{15}(1) (2014)
  1929--1958

\bibitem{krizhevsky2012imagenet}
Krizhevsky, A., Sutskever, I., Hinton, G.E.:
\newblock Imagenet classification with deep convolutional neural networks.
\newblock In: Proceedings of the Advances in Neural Information Processing
  Systems. (2012)  1097--1105

\bibitem{vovk2013kernel}
Vovk, V.:
\newblock Kernel ridge regression.
\newblock In: Empirical inference.
\newblock (2013)  105--116

\bibitem{scikit-learn}
Pedregosa, F., Varoquaux, G., Gramfort, A., Michel, V., Thirion, B., Grisel,
  O., Blondel, M., Prettenhofer, P., Weiss, R., Dubourg, V., Vanderplas, J.,
  Passos, A., Cournapeau, D., Brucher, M., Perrot, M., Duchesnay, E.:
\newblock Scikit-learn: Machine learning in {P}ython.
\newblock Journal of Machine Learning Research \textbf{12} (2011)  2825--2830

\bibitem{chollet2015keras}
Chollet, F.,  et~al.:
\newblock Keras.
\newblock \url{https://keras.io} (2015)

\bibitem{graves2013speech}
Graves, A., Mohamed, A.r., Hinton, G.:
\newblock Speech recognition with deep recurrent neural networks.
\newblock In: Proceedings of the IEEE International Conference on Acoustics,
  Speech and Signal Processing. (2013)  6645--6649

\bibitem{pan2010survey}
Pan, S.J., Yang, Q.,  et~al.:
\newblock A survey on transfer learning.
\newblock IEEE Transactions on Knowledge and Data Engineering \textbf{22}(10)
  (2010)  1345--1359
  
  
\bibitem{passalis}
Passalis, N., Tefas, A.:
\newblock Learning Deep Representations with Probabilistic Knowledge Transfer.
\newblock In: Proceedings of the European Conference on Computer Vision. (2018)  268--284
  

\end{thebibliography}
\end{document}